\documentclass[runningheads]{llncs}

\usepackage[final,year=2024,ID=7779]{eccv}

\usepackage{eccvabbrv}

\usepackage{graphicx}
\usepackage{booktabs}
\usepackage{multirow}
\usepackage{array}
\usepackage{caption}
\usepackage{booktabs} 
\usepackage{setspace}
\usepackage{overpic}

\usepackage[outercaption]{sidecap}
\usepackage[accsupp]{axessibility}  

\newlength\savewidth\newcommand\shline{\noalign{\global\savewidth\arrayrulewidth
  \global\arrayrulewidth 1pt}\hline\noalign{\global\arrayrulewidth\savewidth}}

\usepackage[pagebackref,breaklinks,colorlinks]{hyperref}

\usepackage{orcidlink}

\begin{document}

\title{Instilling Multi-round Thinking into \\ Text-guided Image Editing} 

\titlerunning{Abbreviated paper title}


\author{Lidong Zeng$^{1}$, Zhedong Zheng$^1$, Yinwei Wei$^2$, Tat-seng Chua$^{1}$\\}

\authorrunning{L.~Zeng et al.}

\institute{$^1$ School of Computing, National University of Singapore \\ $^2$ Monash University}

\maketitle

\begin{abstract}
This paper delves into the text-guided image editing task, focusing on modifying a reference image according to user-specified textual feedback to embody specific attributes. Despite recent advancements, a persistent challenge remains that the single-round generation often overlooks crucial details, particularly in the realm of fine-grained changes like shoes or sleeves. This issue compounds over multiple rounds of interaction, severely limiting customization quality.
In an attempt to address this challenge, we introduce a new self-supervised regularization, \ie, multi-round regularization, which is compatible with existing methods. Specifically, the multi-round regularization encourages the model to maintain consistency across different modification orders. 
It builds upon the observation that the modification order generally should not affect the final result.
Different from traditional one-round generation, the mechanism underpinning the proposed method is the error amplification of initially minor inaccuracies in capturing intricate details. 
Qualitative and quantitative experiments affirm that the proposed method achieves high-fidelity editing quality, especially the local modification, in both single-round and multiple-round generation, while also showcasing robust generalization to irregular text inputs. The effectiveness of our semantic alignment with textual feedback is further substantiated by the retrieval improvements on FahisonIQ and Fashion200k.

  
\end{abstract}
\section{Introduction}


\begin{figure*}[t]
	\centering
	\begin{overpic}[width=\textwidth]{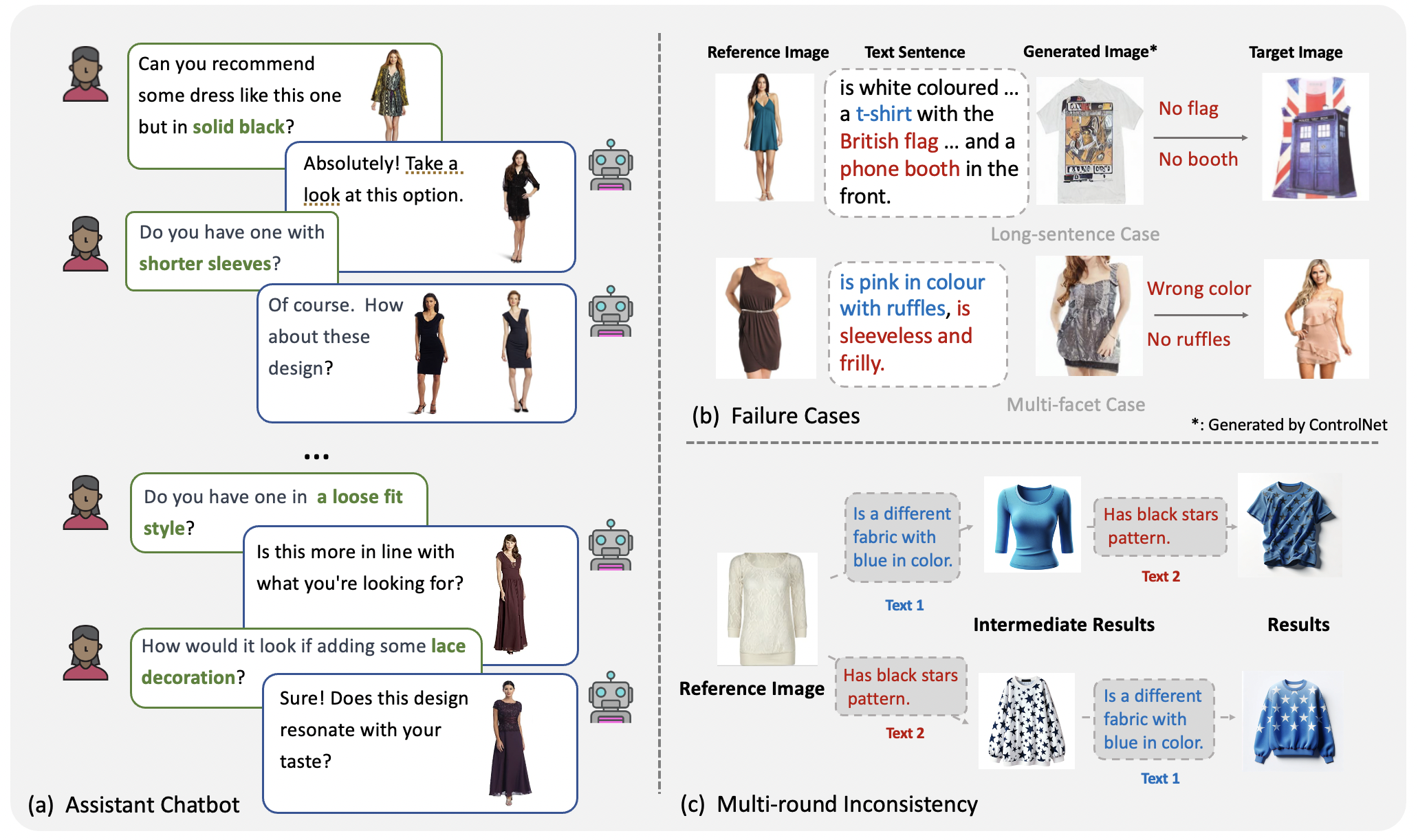}
 \vspace{-.1in}
        \end{overpic}
	\caption{
    (a) A typical use case of multi-round interactive editing. The learned model can understand text instruction and the semantic meaning of images and craft images based on previous user feedback. This real-world scenario often involves multi-round generation rather than single-round generation.
    (b) Some common failure cases on the prevailing methods~\cite{controlnet}, \ie, long sentence ignorance case, multi-facet forgetting case. We could observe the significant visual difference between generated images and ground-truth targets.
    (c) Here we show a typical two-round inconsistency case. The final generated results are sensitive to the order of text guidance.
	} \label{fig:dialogue}
	\vspace{-.1in}
\end{figure*}

%
%
Text-guided image editing is usually regarded as a sub-task of conditional generation to modify the images based on textual cues~\cite{stackgan,attngan,reed2016generative,controlgan}.  
Given a textual description, the generative model is to adjust an existing reference image in accordance with the provided instructions. 
Due to the simple interaction via natural language, such conditional generation has profound commercial potential in applications leveraging user feedback, such as photo editing~\cite{blendeddiffusion, prompt2prompt, glide, dreambooth, textInversion}, customized content creation\cite{liu2021learning, tuneavideo,rodin} and virtual shopping assistant chatbot\cite{fashionchatbot,ye2022reflecting} (see Figure~\ref{fig:dialogue}(a)). 
The key challenge underpinning this task is to ensure 
semantic alignment between text guidance and the generated image.
There are two primary families of existing methods. One line is to apply the Generative Adversarial Networks (GANs)~\cite{gan} via competition between the generator and discriminator. For instance, StackGAN~\cite{stackgan} has achieved high-resolution results by gradually fusing text and image content. 
Another line is the recent diffusion-based generative models~\cite{ddpm}, which iteratively modify the outputs from a random noise. 
The diffusion-based model usually migrates the text feature from advanced pre-trained vision-language model~\cite{clip, raffel2020exploring} 
and facilitates an alignment between textual and visual representations in the latent space during the denoising process. 
For instance, ControlNet~\cite{controlnet}, an extension of Stable-Diffusion~\cite{ldm}, possesses the capability to embed arbitrary image conditions to influence the generated result. 
However, most existing methods, including recent advancements like Stable Diffusion and ControlNet, usually focus on the one-time generation. 
If the generated result is unexpected, users do not have API to provide their feedback to modify the unsatisfied parts, especially failing to capture the representation with long or multi-faceted text description (see Figure~\ref{fig:dialogue}(b)). 
The case is even worse when we delved into these methodologies in the context of multi-round generation to further assess their alignment capabilities, as depicted in Figure~\ref{fig:dialogue}(c). Given that these models are primarily tailored for single-round generation tasks, the error accumulation amplifies throughout each round. Despite the same text guidance, we could observe a relatively large discrepancy between the two generated results. 
In an attempt to address the challenges, our endeavor focuses on enhancing current models by instilling multi-round thinking.
The underpinning motivation for our approach originates from an empirical observation that the order of text is invariant to the quality of generated results. For instance, given a multifaceted text description like ``Make it brighter and have a longer sleeve'', it can be semantically divided into two distinct descriptions: increasing brightness and elongating the sleeve. When generating a result based on such a description in two rounds, the sequence of these instructions should not influence the final outcome.
In contrast, we notice a lack of such consistency in the current model, which is a crucial reason for their inability to maintain coherence with longer sentence comprehension. 
Therefore, we propose a new self-supervised learning that focuses on enhancing generation stability by incorporating multi-round regularization.
In particular, we optimize error accumulation throughout generation rounds and propose a new learning strategy to stabilize the learning process.
Furthermore, by decomposing lengthy sentences into an order-independent sequence of sub-sentences, we also facilitate the model scalability to ill-formed text in real-world scenarios, such as ungrammatical sentences and colloquial expressions. 
In brief, our contributions are as follows:
\begin{itemize}
\item 
We introduce a new self-supervised regularization method seamlessly integrated into existing models. It explicitly motivates the network learning consistency between different modification orders. It largely mitigates error accumulation in the context of multi-round text-guided image generation tasks and also prevents overfitting to only a few keywords in the text during learning.

\item Extensive experiments on two benchmarks, \ie, FashionIQ~\cite{fashioniq} and Fashion200k~\cite{fashion200k}, verify the effectiveness of the proposed method in terms of synthesizing quality, \eg, FID~\cite{fid}, and semantic alignment via retrieval recall rate. 
Albeit simple, the learned model not only improves the generalizability to multi-round synthesize but shows better text alignment even in the single-round generation. 
Besides, the model, considering the broken sentence during learning, also shows great scalability to the ill-formed text understanding. 



\end{itemize}

\section{Related work}

\noindent\textbf{Text-guided Image Generation} Text-guided image generation, as a sub-task of conditional generative tasks, aims to generate an image whose content or style is aligned with the given text description. 
A prevailing approach inherits from the image synthesizing task is to leverage Generative Adversarial Networks (GANs)~\cite{gan}. 
For instance, StackGAN~\cite{stackgan} decomposes cross-modality synthesis into step-by-step sub-problems from low-resolution to high-resolution generation. 
Taking one step further, StackGAN++~\cite{stackgan++} introduces multiple discriminators optimizing different scales, while Zhang \etal~\cite{textasoerator} fuses the object and position attention in the description into the generation process. 
Another line of work adopts the Denoising Diffusion model considering its exceptional capability to produce highly realistic images and training stability
~\cite{DDPMbeatsgan}. Pioneering works, such as DALLE-2~\cite{dalle2}and Imagen~\cite{imagen}, employ a coarse-to-fine approach that crafts images from text description at a reduced resolution and subsequently increases the resolution. 
This strategy mitigates the remaining challenge of high computational resources demand~\cite{understanding} for the diffusion model. Similarly, Stable Diffusion(SD)~\cite{ldm}, an implementation of latent diffusion model, resorts to denoising on a shrink-size latent embedding to address the issue of limited resources. 
Building upon this, Stable Diffusion XL~\cite{sdxl} further scales up the backbone parameters, improving the synthesizing quality.  
Except for the text prompts, some works~\cite{instructp2p, controlnet} attempt to instill additional conditions to control the generating process. 
Instruction Pixel2Pixel~\cite{instructp2p} encodes the input image into the latent space and fuses both image embedding and modification text as the condition.
Furthermore, ControlNet~\cite{controlnet}  facilitates diverse inputs, such as semantic maps, skeletons, and sketches. 
However, these aforementioned methods focus on one-time generation without considering any user feedback. 
In contrast, the proposed method is different from these methods in two aspects: (1) We focus on long-term generation instead of one-time generation, which could largely improve user interaction experience. (2) We observe that the multi-round generation accumulates the error and, in turn, facilitates the fine-grained detail generation. The proposed method is complementary to most existing approaches.

\noindent\textbf{Multi-round Generation Consistency.}
In comparison to single-round generation, multi-round generation strategies often necessitate the continuity of results, where the balance of the modification along rounds and the preservation of historical information is emphasized~\cite{zhou2022tigan,zheng2019joint}. Such a demand is wildly required in vision tracking~\cite{cycleconsistencyoftime}, video generation~\cite{suo2022jointly} and dialogue-based image editing~\cite{tdandp, talktoedit, xia2021tedigan} tasks for the correspondence of previously generated result. 
An early attempt by Wang \etal~\cite{cycleconsistencyoftime} introduces a self-supervised cycle consistency to learn a continuous feature space for video object tracking. 
Suo \etal~\cite{suo2022jointly} presents short-long term consistency in generating sign language teaching videos based on future action to ensure the continuity of generated video, while Zhang \etal~\cite{zhang2023multi} explore the geometry consistency of 3D objects. 
Similarly, recent advancement in the interactive dialogue-like system also demands a history consistency
~\cite{tdandp,seqattngan,textasoerator,talktoedit}, as shown in Figure~\ref{fig:dialogue}(a). 
To this end, 
one line of work introduces a state tracker as historical memory to control generation at each step~\cite{tdandp}, while Chen \etal~\cite{seqattngan} integrates an attention module to leverage the memory.
Another line of work leverages a strong capability of context comprehension and human-like interaction in Large Language Models (LLMs)~\cite{instructgpt,fashionGPT}, 
An update on ChatGPT4 \footnote{\small \url{https://openai.com/chatgpt}} has been released that is capable of both image understanding and iterative image generation combined with DALLE3~\cite{dalle3}. 
Although advanced in dialogue quality, editing with LLMs takes up a large amount of computational resources to train from scratch~\cite{llmoverview}. 
In contrast, our method can be seamlessly integrated into the current framework with efficient fine-tuning to further harness image quality while saving resources. 

\section{Method}


\begin{figure*}[t]
	\centering
	\begin{overpic}[width=\textwidth]{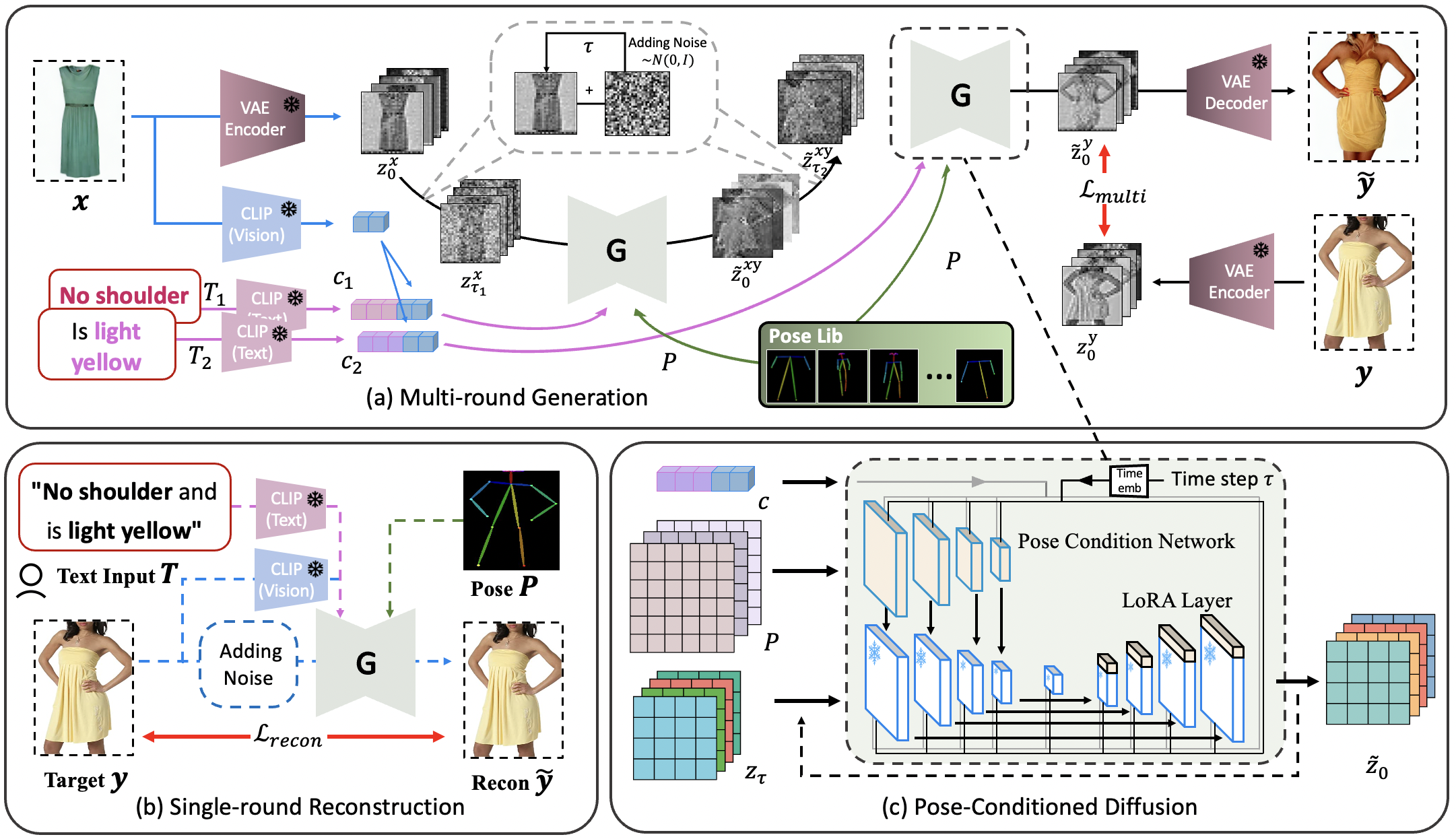}
 \vspace{-.1in}
        \end{overpic}
	\caption{
            A schematic overview of our framework.
            (a) \textbf{Multi-round Generation:}
            The multi-round regularization is achieved by a skip loss only supervising on final output $\tilde{z}^y_0$ and ground truth $z^y_0$. Starting with the encoding of the reference image $x$ into a latent embedding $z^x_0$, we conduct a complete denoising process twice to get $\tilde{z}^y_0$. The information of $x$ is traced by the blue line, while the pink line indicates the flow of text information.
            (b) \textbf{Single-round Reconstruction:}
            In single-round reconstruction, the objective is to reconstruct the target image $y$ by denoising on perturbed ground truth $y$ alongside the corresponding text condition $T$ and pose $P$. (c) \textbf{A brief illustration of Pose-Conditioned Diffusion.} Given the input $c$, $P$ and $z_\tau$, the diffusion is to generate $\tilde{z}_0$ via iteratively denosing on the time step $\tau$. 
            We adopt an extra LoRA layer, which is concatenated into each attention block of the U-net decoder.
	} \label{fig:overview}
	\vspace{-.1in}
\end{figure*}

\noindent\textbf{Problem Definition.}
Given a triplet input, \ie, a text sentence $T$, reference image $x$, 
we intend to learn a generative model to craft target image $y$, which is semantically modified $x$ according to $T$. 
As shown in Figure~\ref{fig:overview}, our framework consists of three main parts, text encoder, image encoder, and Diffusion generator $G$ with Variational AutoEncoder(VAE)~\cite{vae} encoder and decoder.
Without loss of generalization, we follow the previous works~\cite{ldm,DDPMbeatsgan} to deploy an off-the-shelf CLIP model as our vision and text encoder, and we exploit the pretrained Stable Diffusion (SD)~\cite{ldm} as the backbone of our Diffusion generator $G$. 
Since the output image can be in arbitrary poses, to facilitate the deterministic reconstruction loss, we also introduce the pose condition $P$. 
During training, $P$ is the ground-truth pose extracted from target image $y$ by~\cite{fang2022alphapose}. When inference, the target pose condition can be randomly sampled from the pose library. 

\subsection{Multi-round Learning} 
During multi-round learning, we mainly focus on the two-round consistency (see Figure~\ref{fig:overview}(a)). We split the original text description $T$ as two partial texts $T_1$ and $T_2$. 
Given $T_1$, $T_2$, and the reference image $x$, we extract and concatenate the high-level text and visual embedding as $c_1$ and $c_2$ respectively. 
Besides, we also extract the low-level feature map $z^x_0$ via VAE encoder and add Gaussian noise as $z^x_{\tau_1}$, where we use subscript to indicate the noise time step $\tau$. During training, $\tau$ is randomly selected in [1, 1000]. 
Then we adopt the Diffusion generator $G$ to denoise on $z^x_{\tau_1}$ for iteration of $\tau_1$ under the conditions of semantic embedding $c_1$ and $P$. 
As shown in Figure~\ref{fig:overview}(c), $G$ contains skip connect U-net with $4$-block encoder and $4$-block decoder. We also introduce an extra pose condition network to encode the conditions $P$. 
The pose condition network has the same encoder structure as the main U-net branch. The layer-wise activation of the pose condition network is added to the corresponding encoder layers of the U-net. 
Except for the pose embedding, the high-level semantic condition $c$ and time embedding from $\tau$ are also inserted into both U-net blocks and the pose condition network. 
Besides, we enable the learnable parameters in the decoder of U-net by inserting several Low-rank adaption (LoRA) layers.
During the first denoising loop, $G$ predicts the intermediate result $\tilde{z}^{xy}_0$ as:
\begin{equation}
    \tilde{z}^{xy}_0 = G^{\tau_1}_0(z^x_{\tau_1},c_1,P), 
\end{equation}
which is aligned with the text embedding $c_1$ and pose $P$. The superscript and subscript of $G$ denote the diffusion start and end timestep respectively. \textbf{Here we do not involve the iterative diffusion process for illustration simplicity.}  
Then we apply the noise process and adopt the diffusion model $G$ again with the rest text condition $c_2$ to obtain the final $\tilde{z}^{y}_0$. 
It is worth noting that we adopt the independent Gaussian distribution, and the noise steps $\tau_1$ and $\tau_2$ are different in two rounds.
Considering both encoder and decoder are fixed and off-the-shelf, we could directly apply the feature-level generation loss. Therefore, the multi-round generation loss can be formulated as:
\begin{align}
      \mathcal{L}_{multi} &= ||z^y_0- \tilde{z}^y_0|| = ||z^y_0-G^{\tau_2}_0(\tilde{z}^{xy}_{\tau_2},c_2,P)|| \\
      &= ||z^y_0-G^{\tau_2}_0(G^{\tau_1}_0(z^x_{\tau_1},c_1,P)+\mathcal{N}(\tau_2),c_2,P)||. 
\end{align}
where we could derive the loss backpropagation on $G$ twice. The ground-truth latent $z^y_0$ is extracted from target image $y$, to provide reconstruction supervision on $\tilde{z}^y_0$. 
$\mathcal{N}(\tau)$ is the Gaussian noise which is conditioned on $\tau$. In practice, we usually could simplify this loss, by considering one-step reconstruction instead of recovering to the $0$-th timestep as: 
\begin{equation}
    \mathcal{L}_{multi} = ||z^y_{\tau_2-1}-G^{\tau_2}_{\tau_2-1}(\tilde{z}^{xy}_{\tau_2},c_2,P)||.
\end{equation}
This loss focuses on reconstructing the last time step $z^y_{\tau_2-1}$ according to the latent feature map $z^{xy}_{\tau_2}$, which eases the optimization difficulty. It is worth noting that this loss gradient is also propagated back to the first round via $\tilde{z}^{xy}$.

\noindent\textbf{Discussion.} 
\textbf{Why multi-round property is necessary?} By using the proposed multi-round generation mechanism, we explicitly deploy the diffusion generator twice. It helps to propagate the loss and prevent overfitting. Compared to the single-round generation in most existing works, some local generation artifacts are usually ignored due to the global pattern containing a larger area. 
In contrast, the proposed multi-round generation, due to the multi-round error accumulation, has a stronger motivation to deal with local cases. This property is verified in the experiment. The proposed method has achieved a higher semantic alignment score than widely-used single-round generative approaches, yielding better local generation quality, such as dress ruffles and buttons in Figure~\ref{fig:qualitative}.


\subsection{Single-round Learning}
For conventional single-round learning, we deploy conditional generation and unconditional self-reconstruction to regularize the whole training process. 

\noindent\textbf{Single-round Generation.} 
Compared to the multi-round generation, we also adopt the vanilla single-round generation with the complete sentence $T$. The process is similar to multi-round generation, while we only apply the denoising diffusion $G$ once. Similarly, given the target image $y$,  we first employ a pre-trained vision and text model to extract high-level semantic condition $c$. 
Then we integrate the pose $P$ into the Diffusion module $G$ to reconstruct the noise map. The loss can be formulated as:
\begin{equation}
    \mathcal{L}_{single} = ||z^y_{\tau-1}-G^\tau_{\tau-1}(z^x_\tau,c,P)||. 
\end{equation}
Here we skip the intermediate result $z^{xy}$. Therefore, the gradient is to update the $G$ once to reconstruct the target $z^y_{\tau-1}$ via $z^x_\tau$ and conditions.

\noindent\textbf{Single-round Self-reconstruction.} Depicted in Figure~\ref{fig:overview}(b), our framework also adopts the vanilla self-reconstruction to regularize the training process. 
For the self-reconstruction task, we do not require text embedding as guidance for self-reconstruction, so here we set $c$ as a zero embedding $\emptyset_c$. 
During practice, we adopt the feature reconstruction as:  
\begin{equation}
    \mathcal{L}_{recon} = ||z^y_{\tau-1}- G^\tau_{\tau-1}(z^y_\tau, \emptyset_c, P)||. 
\end{equation}

\noindent\textbf{Discussion.} \textbf{Why still keep single-round learning?} 
Single-round training serves as a good regularization for multi-round learning. The model is easy to collapse if we only adopt the multi-round loss. It is also due to the error accumulation, especially when failing to generate the meaningful intermediate $z^{xy}$. Therefore, we design the optimization strategy in the next subsection to leverage both single-round and multi-round learning, eventually improving the final model generalizability. 

\subsection{Optimization}

As mentioned before, we jointly penalize generator $G$ on single-round and multi-round generation by composing objectives, which can be formulated as:
\begin{equation}
    \mathcal{L}_{total} = \mathcal{L}_{single}+\mathcal{L}_{recon} + \lambda\cdot\mathcal{L}_{multi},
\end{equation}
where $\lambda$ is a dynamic hyper-parameter controlling impact of $\mathcal{L}_{multi}$. 
$\lambda$ gradually decreases from 1 to 0 during training, considering that the final task is to compare the single-round generation. Therefore, we have to gradually shift the focus to the single-round loss. 
When $\lambda = 1$, we focus on both the single-round and multi-round generation $\mathcal{L}_{multi}$, which establishes a long-term connection for semantic alignment. 
When $\lambda = 0$, the model will pay attention to the single-round generation $\mathcal{L}_{single}$ and reconstruction $\mathcal{L}_{recon}$, which regresses to the conventional approaches. 
Considering the trade-off between the multi-round and single-round generation quality, we adopt both kinds of learning at the beginning, and then gradually shift to the single-round generation. 
In particular, we linearly decline the weight $\lambda$. More ablation studies can be found in Section~\ref{sec:lambda}. 

\section{Experiment}


\subsection{Datasets and Evaluation Metrics}
We mainly verify the effectiveness of the proposed method on two multi-modal retrieval datasets, \ie, FashionIQ~\cite{fashioniq} and Fashion200k~\cite{fashion200k}. 
\textbf{FashionIQ}~\cite{fashioniq} involves over 17.5k queries for training and 5.5k for evaluating and each query consists of a candidate image, text user feedback, and target image. Considering the dominance of the model pose that appeared in \textit{dress} subset, we intend to instill pose information to the ControlNet for pose disentangling. 
\textbf{Fashion200k}~\cite{fashion200k} consists of over 200k fashion images with their corresponding text caption. To extract the triplets query, we follow the previous work~\cite{retrievalwithfeedback} that manually constructs triplets by comparing the difference between two arbitrary image-caption pairs. In addition, considering the images in Fashion200k usually contain only part of the body, we deploy a slightly different condition, \ie, coarse-grained canny edge, to control the network.  

\noindent\textbf{Evaluation Metrics.} We evaluate our method from both image quality and the alignment of semantic meaning. \textbf{(1)} The visual quality is assessed by Fréchet Inception Distance (FID)~\cite{fid}. 
\textbf{(2)} The semantic alignment between user feedback and generated image is measured by image recall rate and the average CILP score~\cite{clip} over generated images and target images.
For the retrieval dataset , we also introduce recall rate, \ie, Recall@K, as a semantic metrics to measure the text-image alignment after the image editing. Recall@K indicate whether the ground-truth image is in the top-K of retrieval list. We adopt the fine-tuned ResNet-based~\cite{resnet} image encoder to extract the visual feature for retrieval, following the open-source code \cite{retrievalwithfeedback}.
\begin{figure*}[t]
	\centering
 \vspace{-.0in}
	\begin{overpic}[width=\textwidth]{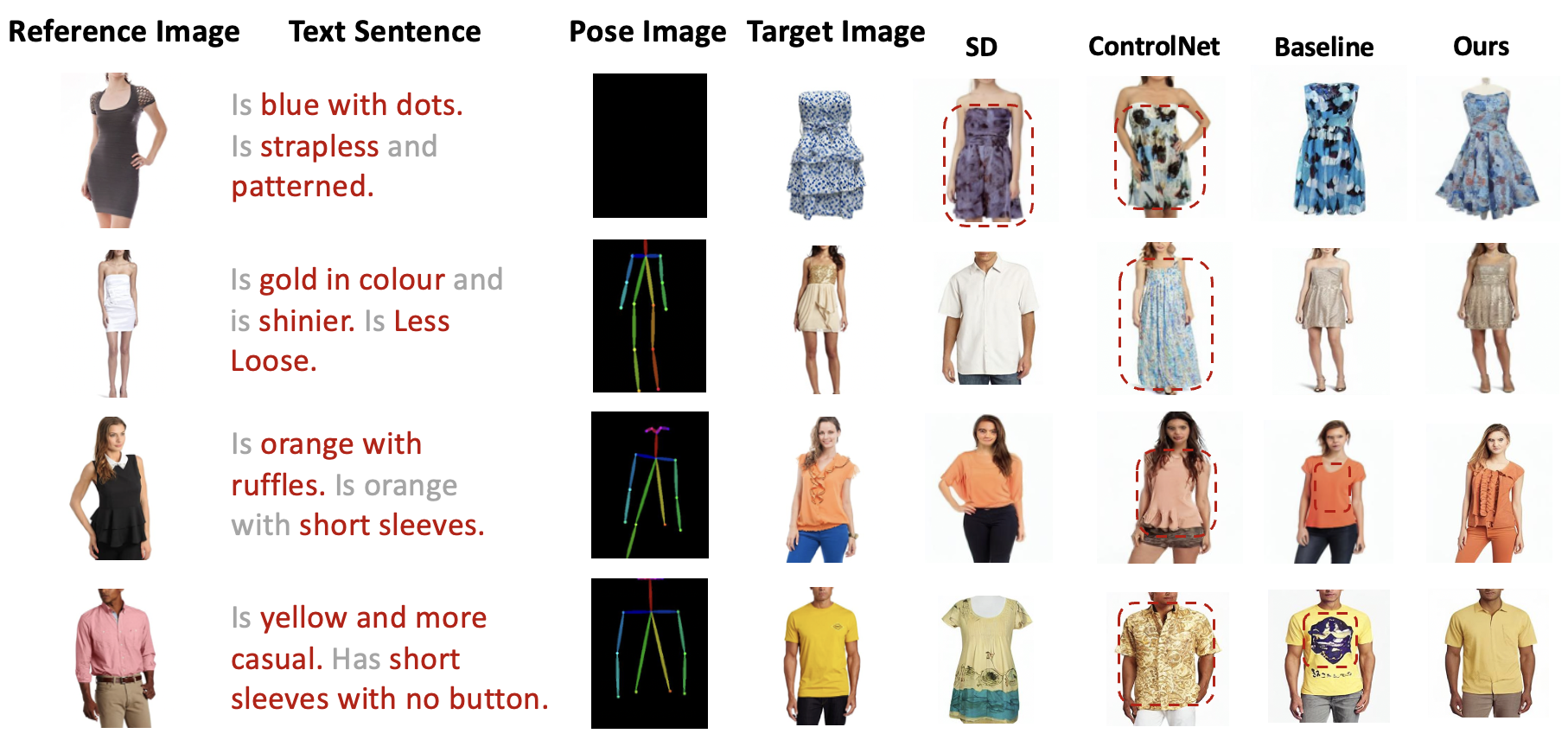}

        \end{overpic}
        \vspace{-.25in}
	\caption{
            Comparison between Stable Diffusion (SD), ControlNet, baseline, and our proposed method in the single-round generation. The \textcolor{red}{red dashed box} indicates the mismatch areas between the generated results and the corresponding text sentences, respectively. The baseline shares the same structure and settings except for multi-round loss $\mathcal{L}_{multi}$. We could observe the baseline sometimes miss keywords, \ie, ``ruffles'' (the 3rd row), and over-modify ``casual'' in the (the 4th row), while the proposed method could notice such descriptive words.
	} \hspace{-3mm}\label{fig:qualitative}
	\vspace{-.1in}
\end{figure*}

\subsection{Implementation Details}
We implement our approach in PyTorch~\cite{pytorch}. We deploy AdamW~\cite{adamw} to train $G$ with learning rate to $2e^{-4}$ and $(\beta_1 , \beta_2) =(0.95, 0.99)$. We also adopt the learning rate warm-up with $0.1$ at the beginning. 
As the diffusion process contains stochastic noise, if the intermediate results $z^{xy}$ are not good at the beginning, the second round generation will be largely impacted and lead to model collapse.
The learning rate scheduler is set as cosine decay. We train the model for 200 epochs.
The input images are uniformly resized to $256 \times256$ and then normalized between $[-1,1]$, processed by VAEencoder to obtain the latent. 
Similarly, Pose images are uniformed to $[0,1]$ with the size of $256 \times 256$.
Meanwhile, a copy of the input image is resized to $224 \times 224$ for extracting high-level image features via the CLIP vision model. 
We truncate the text with the upper length of 30, and then padding the short sentence to 30. 
Therefore, we compose the text embedding and image embedding as the condition, which is with the shape of $ (30+50) \times 768$.
In the multi-round generation, the text $T$ is split according to the meaning. 
In particular, 
we extract the difference between the reference image caption and the target image caption as the input text, while FashionIQ provides a ready-made split.
We modified the U-net by inserting LoRA layers into its decoder block. Specifically, we add LoRA to each $K, Q, V$ layer in the cross-attention module, with LoRA rank $4$.

\noindent\textbf{Multi-round optimization.} During the multi-round training session, we utilize the reparameterization in \cite{understanding} to generate the first-round image in one denoising step instead of multiple steps. Therefore, we could largely save the training time. The total training time is about 15 hours on the FashionIQ dataset with one Nvidia V100 GPU. 

\noindent\textbf{Reproducibility.} We will release Pytorch code for re-implement all results.

\begin{SCfigure*}[][h]
	\centering
  \vspace{-.1in}
	\includegraphics[width=0.43\columnwidth]{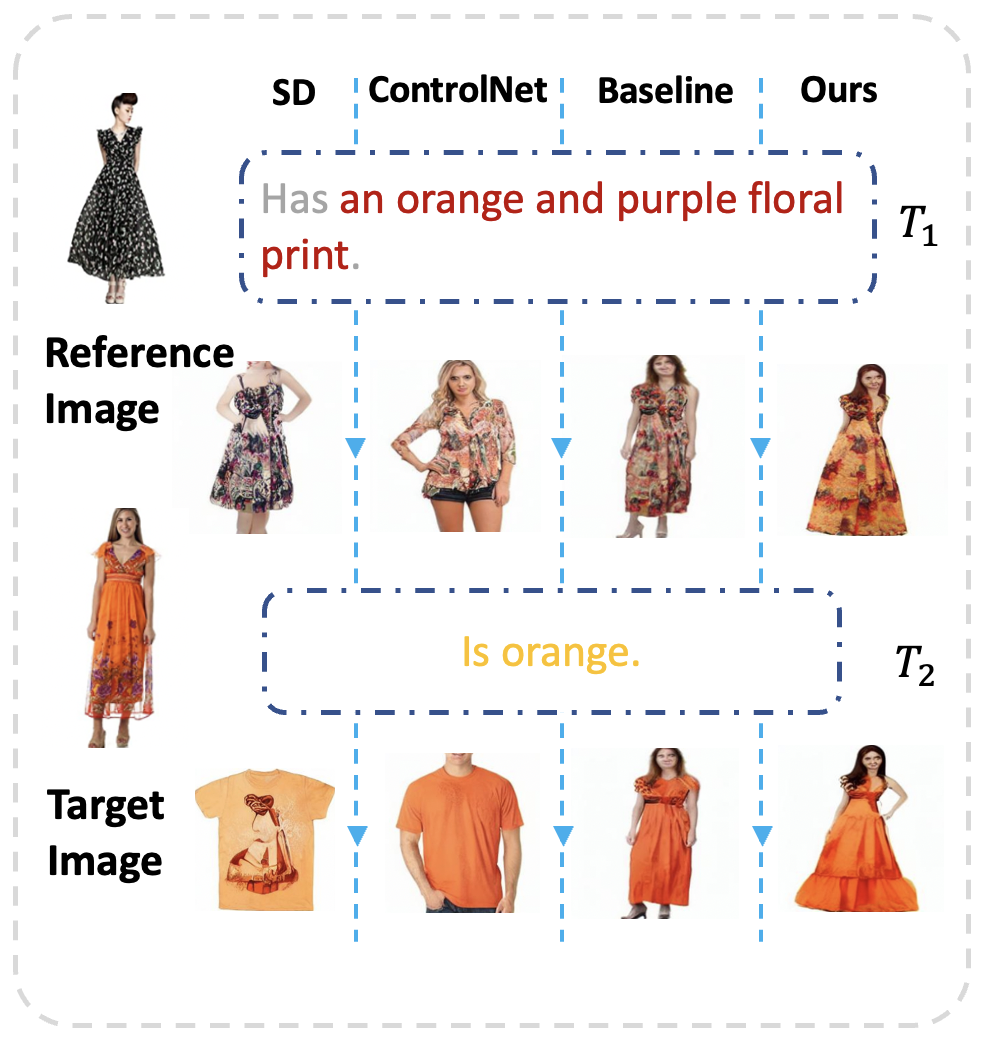}
	\caption{
            Qualitative comparison between Stable Diffusion, ControlNet, our baseline model without multi-round constraint, and our proposed model. Starting from the reference image, each generation is based on previous results. 
            We could observe that our method generates reasonable results, while better preserving the style of reference image.
	} \label{fig:qualitative_multiround}
\end{SCfigure*}

\subsection{Comparison with SOTA}


\noindent\textbf{Qualitative Comparison.} We first qualitatively compare our method with two common generative models, \eg, ControlNet and Stable Diffusion (SD). 
In the single-round generation, as shown in Figure~\ref{fig:qualitative}, the misalignment between text and the generated image is highlighted by a dashed box with the corresponding color. 
We observe three primary points. \textbf{(1)} The Stable Diffusion and ControlNet have more failure cases in capturing the detailed text meaning, such as editing sleeve length and changing the color depicted.
\textbf{(2)} 
The main difference between the baseline and ControlNet is the semantic condition from the reference image. 
The enhancement in image generation and local editing is mainly due to the better preservation of semantic patterns by our condition fusion.
\textbf{(3)} Besides, we could observe that the baseline model is also imperfect. For instance, the baseline misses keywords like ``ruffles'' with flat texture, while being sensitive to ``casual'' with an undefined logo. In contrast, the proposed method with the multi-round loss regularizes the training and shows a more consistent generation quality. 
On the other hand, we also evaluate these models in a multi-round generation by splitting the text as we elucidated in our methodology. 
Compared to the single-round generation, the split text is more challenging since it is more sparse in meaning. 
As shown in  Figure~\ref{fig:qualitative_multiround}, Stable Diffusion and ControlNet lost the original reference information after two rounds of generation. Both models change the dress to the common category, \ie, a T-shirt. 
In contrast, the baseline model performs well on ``dress'', but it still misses the fine-grained ``crinoline dress'' pattern from the reference image. The main reason is that the intermediate results of the three methods, \ie, Stable Diffusion, ContorlNet and baseline already miss some key patterns.
Compared with these methods,
our multi-round regularized model mainly changes color and texture in the second round of generation, maintaining consistency in local editing. We also could observe that the output result is controllable as it aligns well with the text modification.

\begin{table}[t]
	\footnotesize
	\renewcommand\tabcolsep{3pt}
 \caption{Quantitative comparison on image generation with the FashionIQ and the Fashion200k dataset.}\label{tab:FIDandCLIP} \vspace{-.1in}
	\centering
 \resizebox{0.95\textwidth}{!}{
\begin{tabular}{ccccccc}
\shline \small
\multirow{2}{*}{Dataset}   & \multicolumn{2}{c}{Stable Diffusion~\cite{ldm}}  

& \multicolumn{2}{c}{ControlNet\cite{controlnet}}            
& \multicolumn{2}{c}{Ours} 
\\ \cline{2-7} 
& FID $\downarrow$             
& CLIP Score $\uparrow$        
& FID $\downarrow$              
& CLIP Score $\uparrow$            
& FID $\downarrow$   
& CLIP Score $\uparrow$            
\\ \hline
\multicolumn{1}{c|}{FashionIQ} & 
9.26 & \multicolumn{1}{c|}{0.64}  & 
8.56 & \multicolumn{1}{c|}{0.66} & 
8.58 & \multicolumn{1}{c}{0.71} \\
\multicolumn{1}{c|}{Fashion200k} & 
8.11  &  \multicolumn{1}{c|}{0.72} & 
6.83  & \multicolumn{1}{c|}{0.80} & 
6.54 & \multicolumn{1}{c}{0.81} \\

 \shline
\end{tabular}}
\vspace{-2mm}
\end{table}
\begin{table}[t]
\caption{
		Quantitative comparison between Stable Diffusion, ControlNet, and our method on the FashionIQ, Fashion200k dataset in terms of retrieval. Overall denotes that we deploy all test categories as the retrieval gallery. 
	}\label{tab:recall}
	\vspace{-1mm}
	\centering
\resizebox{\linewidth}{!}{
\begin{tabular}{cccccccc}
\shline
\multicolumn{2}{c}{Dataset}   & \multicolumn{2}{c}{Stable Diffusion} &\multicolumn{2}{c}{ControlNet} & \multicolumn{2}{c}{Ours} 
\\ \cline{3-8}
~ &~ 
&Recall@10 $\uparrow$
&Recall@50 $\uparrow$
&Recall@10 $\uparrow$
&Recall@50 $\uparrow$
&Recall@10 $\uparrow$
&Recall@50 $\uparrow$
\\ \hline
\multirow{4}{*}{FashionIQ} & \multicolumn{1}{|c|}{\textit{dress}} & 
7.17 &\multicolumn{1}{c|}{20.59} &
21.34 &\multicolumn{1}{c|}{47.33}& 
34.17 &59.12 \\ 
& \multicolumn{1}{|c|}{\textit{shirt}} & 
5.08&\multicolumn{1}{c|}{16.09} & 
23.56 &\multicolumn{1}{c|}{46.44}  & 
28.56 &54.34 \\ 
& \multicolumn{1}{|c|}{\textit{toptee}} &
8.15 &\multicolumn{1}{c|}{19.87} &
32.45 &\multicolumn{1}{c|}{59.77}  & 
39.92 &68.40 \\ \cline{2-8}
& \multicolumn{1}{|c|}{\textit{overall}} &
4.96 &\multicolumn{1}{c|}{13.31} &
22.41 &\multicolumn{1}{c|}{44.52}  & 
\textbf{32.07} &\textbf{56.10}\\  \hline
                            
\multirow{6}{*}{Fashion200k} & \multicolumn{1}{|c|}{\textit{dress}} &
0.67 &\multicolumn{1}{c|}{2.49} &
31.10 &\multicolumn{1}{c|}{45.39}  &
36.18 &50.58 \\ 
& \multicolumn{1}{|c|}{\textit{jacket}} & 
0.68 &\multicolumn{1}{c|}{2.47} &
17.92 &\multicolumn{1}{c|}{32.31}  &
20.39 &36.31 \\ 
& \multicolumn{1}{|c|}{\textit{pants}} &
0.70 &\multicolumn{1}{c|}{2.79} &
15.31 &\multicolumn{1}{c|}{29.10}  &
18.24 &33.48\\ 
& \multicolumn{1}{|c|}{\textit{skirt}} &
0.62 &\multicolumn{1}{c|}{2.74} &
23.94 &\multicolumn{1}{c|}{37.69}  & 
28.43 &43.55 \\ 
& \multicolumn{1}{|c|}{\textit{top}} & 
0.56 &\multicolumn{1}{c|}{2.27} &
23.18 &\multicolumn{1}{c|}{37.63}  & 
27.44 &42.59 \\ \cline{2-8}
& \multicolumn{1}{|c|}{\textit{overall}} &
0.19 &\multicolumn{1}{c|}{0.92} &
17.47 &\multicolumn{1}{c|}{27.96}  & 
\textbf{21.34} &\textbf{32.39} \\ 
\shline


\end{tabular}
}
\end{table}
\noindent\textbf{Quantitative Comparison.} We study the generation quality and semantic alignment in this part, consolidating our observation from the qualitative evaluation. 
In particular, we train and evaluate models on FashionIQ and Fashion200k with FID score on generation fidelity, while the CLIP score presents semantic alignment. 
Firstly we compare models on one round generation with full text shown in Table~\ref{tab:FIDandCLIP}. Our model consistently outperforms the ControlNet and Diffusion models on both metrics. Specifically, our model achieves the CLIP Score of 0.71 on FashionIQ, outperforming Stable Diffusion and ControlNet, which scored 0.64 and 0.66, respectively. Then we evaluate models on consecutive two rounds setting with accumulated text in Table~\ref{tab:mmtable}. Our model outperforms others by at least $7\%$ on CLIP Score with competitive FID Score.
Considering the generation quality on FID, the proposed method attains a competitive generation quality, while better aligning with the semantic meaning. 
Furthermore, we evaluate our method by content-based image retrieval metrics (see Table~\ref{tab:recall}). Specifically, by using the model-generated image to search for the ground-truth image, we evaluate the semantic meaning of the synthesizing quality at a fine-grained level. 
We follow the existing method~\cite{retrievalwithfeedback} to train the models on two datasets respectively, utilizing the trained image encoder to extract features for retrieval. 
We find that 
the proposed model has achieved $32.07\%$ Recall@10 and $56.10\%$ Recall@50 accuracy on FashionIQ. It significantly surpasses ControlNet by $+9.66\%$ Recall@10, $+10.58\%$ Recall@50, even in the sub-categories. On Fashion200k, we observe a similar margin between the proposed method and Stable Diffusion as well as ControlNet. 
It verifies the effectiveness of the proposed method on semantic alignment after editing.

\begin{table}[t]
\caption{Ablation studies on the FashionIQ dataset. 
    (a) Ablation study on the effectiveness of the component and settings to our model. Models trained and evaluated in terms of FID and retrieval rate on FashionIQ. (b) Comparison of model performance under ill-text conditions. The performance is evaluated by image similarity extracted by the CLIP image encoder as CLIP score (higher is better). The similarity is compared between the generated image and the target image for all cases except $Swap^*$. $Swap^*$ consistency is calculated between two-step generation results with different sentence orders as Figure~\ref{fig:dialogue}(c). (c) Quantitative of two-round results on FashionIQ. The 2 rounds text condition is split from one sentence, therefore the performance is evaluated on the 2nd round generation as the intermediate ground truth is infeasible. (d) Ablation study on the choice of $\lambda$ on FashionIQ. $1 \rightarrow 0$ denotes linearly decreasing $\lambda$, while $0 \rightarrow 1$ gradually increase the value. Baseline arrives good generation fidelity (FID), while missing the editing demands with poor recall rates. In contrast, the proposed method significantly improves the editing semantic alignment. }
\vspace{-.15in}
\begin{subtable}{.6\linewidth}
\centering \small
\caption{}
\label{tab:ablation}
\resizebox{0.9\linewidth}{!}{
\begin{tabular}{l|c|c|c|c|c|c} 
\shline 
Sampling  & $\mathcal{L}_{recon}$ & Init.$x$ & $\mathcal{L}_{multi}$ & R@10$\uparrow$& R@50 $\uparrow$& FID $\downarrow$  \\ \hline
DDIM~\cite{ddim} & \checkmark& &  &   
21.67   &     42.33   & 17.90        \\
DDIM~\cite{ddim} &    \checkmark  & \checkmark   &       &   22.76    &   42.70     & 8.58                     \\
DDIM~\cite{ddim} & \checkmark & \checkmark & \checkmark &31.59 & 55.35& 9.75\\
\hline
DDPM~\cite{ddpm} & \checkmark  & \checkmark   &         &  27.07      &     47.98   & \textbf{7.46}         \\
DDPM~\cite{ddpm}   & \checkmark   & \checkmark   & \checkmark       & \textbf{32.07}      & \textbf{56.08}       & 8.58                     \\ \shline
\end{tabular}
}  \vspace{.05in}
\hfill
\centering 
    \caption{} \vspace{-.1in} \scriptsize
    \resizebox{0.9\linewidth}{!}{
    \label{tab:swap}
\begin{tabular}{lcccccccc}
\shline
Case    & \multicolumn{2}{|c}{Stable Diffusion} & \multicolumn{2}{|c}{ControlNet} &\multicolumn{2}{|c}{Baseline}   & \multicolumn{2}{|c}{Ours}   \\ \hline
\textit{Swap$^*$}  & \multicolumn{2}{|c}{0.772}   & \multicolumn{2}{|c}{0.754}  &\multicolumn{2}{|c}{0.837}  & \multicolumn{2}{|c}{\textbf{0.882}}   \\ \hline

\textit{Rotate}  & \multicolumn{2}{|c}{0.646}   & \multicolumn{2}{|c}{0.667}  &\multicolumn{2}{|c}{0.687}  & \multicolumn{2}{|c}{\textbf{0.703}}   \\ \hline

\textit{Mask n=2}  & \multicolumn{2}{|c}{0.656}   & \multicolumn{2}{|c}{0.677}  &\multicolumn{2}{|c}{0.686}  & \multicolumn{2}{|c}{\textbf{0.707}}   \\ \hline

\textit{Mask n=5}  & \multicolumn{2}{|c}{0.647}   & \multicolumn{2}{|c}{0.677}  &\multicolumn{2}{|c}{0.678}  & \multicolumn{2}{|c}{\textbf{0.697}}   \\ \hline

\textit{Mask n=9} & \multicolumn{2}{|c}{0.623}   & \multicolumn{2}{|c}{0.658}  &\multicolumn{2}{|c}{0.669}  & \multicolumn{2}{|c}{\textbf{0.692}}   \\ \shline

\end{tabular}
    
}
\end{subtable}   
\begin{subtable}{.4\linewidth}
\centering 
\caption{} \vspace{-.1in}
\small
\label{tab:mmtable}\resizebox{0.95\linewidth}{!}{
    \begin{tabular}{lcc}
\shline
Methods   & \multicolumn{2}{|c}{Two-Round}  
\\ \hline

 
 \textit{SD}  &\multicolumn{1}{|c}{12.42 FID $\downarrow$}  & \multicolumn{1}{|c}{0.6533 CLIP $\uparrow$}   \\ \hline

 \textit{ControlNet}  &\multicolumn{1}{|c}{14.71 FID $\downarrow$}  & \multicolumn{1}{|c}{0.6409 CLIP $\uparrow$}   \\ \hline

 \textit{Baseline}  &\multicolumn{1}{|c}{11.95 FID $\downarrow$}  & \multicolumn{1}{|c}{0.6860 CLIP $\uparrow$}   \\ \hline

 \textit{Ours}  &\multicolumn{1}{|c}{13.22 FID $\downarrow$}  & \multicolumn{1}{|c}{0.7006 CLIP $\uparrow$}   \\ \shline

\end{tabular}
} \vspace{.05in}
\hfill
\centering
\caption{} \vspace{-.1in}
\scriptsize
\label{tab:lambda_m} 
    \begin{tabular}{l|c|c|c} 
    \shline
    $\lambda$ &  R@10$\uparrow$& R@50 $\uparrow$& FID $\downarrow$  \\ \hline
    fix $1$ &    30.73 &  54.70 &  16.45   \\
    $0 \rightarrow 1$ &    \textbf{32.10}  &  54.46    &  14.89  \\
    $1 \rightarrow 0$ & 32.07 & \textbf{56.08} & 8.58 \\
    Baseline (fix $0$) &   27.07 & 47.98 & \textbf{7.46} \\ \shline
    \end{tabular}
\end{subtable}
\vspace{-.1in}
\end{table}

\subsection{Ablation Studies and Further Analysis} 

\noindent\textbf{Discussion on Inference Tricks.}
As shown in Table~\ref{tab:ablation}, we study several tricks to ensure the best inference setting. First, we mainly study the DDIM~\cite{ddim} and DDPM~\cite{ddpm} scheduler. We find the DDPM with Gaussian noise performs better in our case since we usually need to significantly modify the image patterns.  
Second, we try to initialize the noise based on the reference image $x$, instead of Gaussian Noise from scratch. We find this trick also could improve the image quality slightly. It is reasonable since the reference image could provide some position information to help recover the noise.

\noindent\textbf{Effect of Multi-round Loss.} 
In Table~\ref{tab:ablation}, we also study the effect of the multi-round loss on the FashionIQ dataset. The proposed method has achieved $32.07\%$ Recall@10, and $56.08$ Recall@50 accuracy while maintaining a competitive FID of $8.56$. Compared with the baseline model without the multi-round loss $\mathcal{L}_{multi}$, the proposed method significantly shows a better fine-grained generation by surpassing $+8.10\%$ Recall@50. 
We also notice that the baseline method has achieved a slightly better FID of 7.46. It is within the expectation since the multi-round constraint is devised intuitively to prevent a toxic deviation in multiple rounds and improve consistency.  For the single-round generation, the proposed method is still competitive but is not better than the end-to-end tuned baseline.

\begin{figure}[t]
	\centering
	\includegraphics[width=0.9\linewidth]{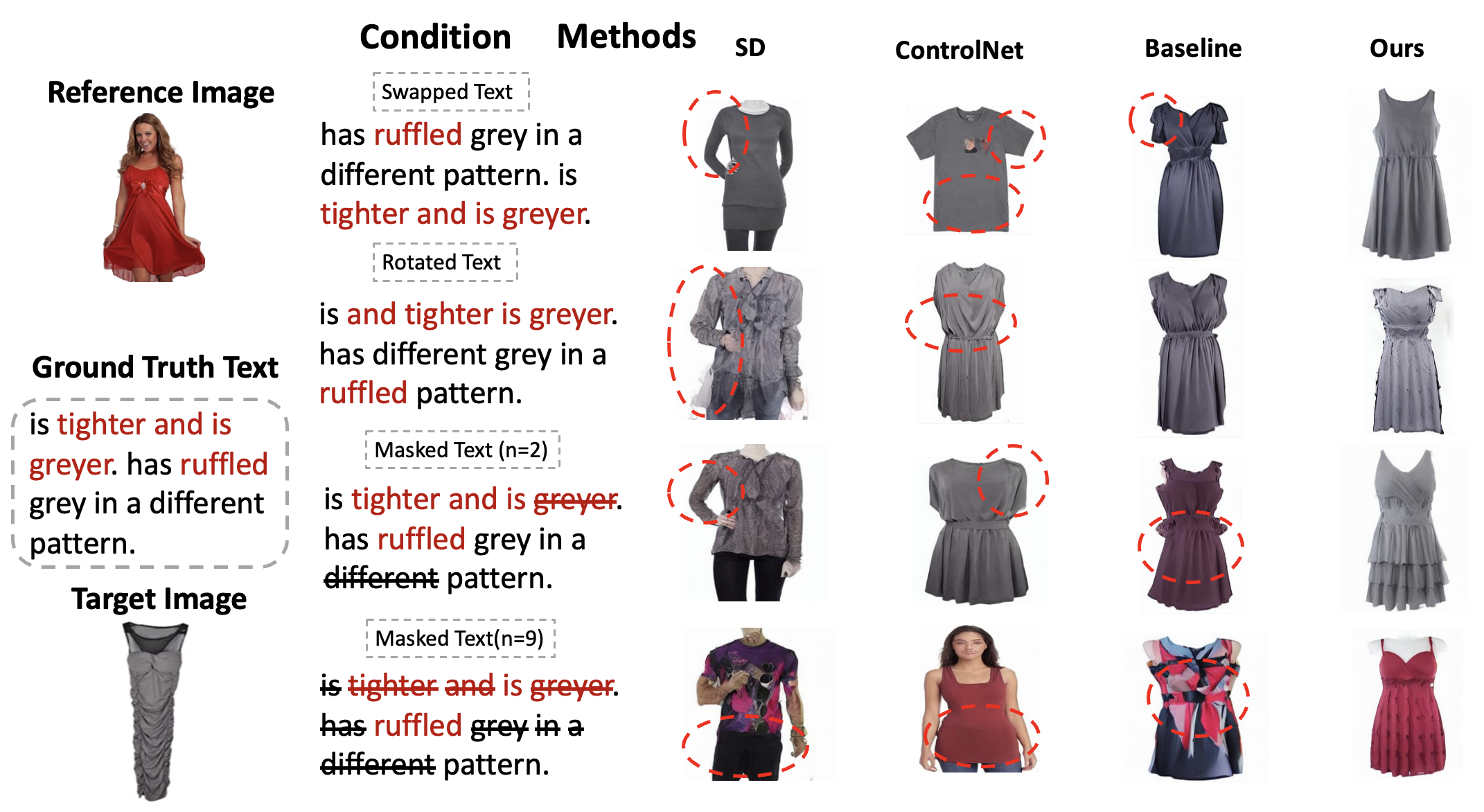}
        \vspace{-.1in}
	\caption{
    Visualization of generation under ill-formed text condition corresponding to Table~\ref{tab:swap}. 
    Comparing to Stable Diffusion, ControlNet and our baseline, the proposed method is robust to different ill-formed texts, \eg, swapping the sentence order, rotating the word order, and masking words. The output is consistent with the corresponding text, \eg, ``tighter'' and ``ruffled''.  
	} \label{fig:ill-text}
        \vspace{-.2in}
\end{figure}

\noindent\textbf{Robustness against Ill-formed Text.}
To further explore the robustness of the proposed model against ill-formed text, we compare our full model with the baseline, ControlNet, and Stable Diffusion for swapping text, mask words, and rotate words experiments. As shown in Table~\ref{tab:swap} and visualized in Figure~\ref{fig:ill-text}, the model performance is mainly evaluated by CLIP score on FashionIQ. 
\textbf{(1)} We first study generation consistency to the sentence order in the multi-round generation. 
Specifically, we split the text $T$ into $T_1, T_2$.  
We demand the model to generate two images for $\{T_1, T_2\}$ and the swapped order $\{T_2, T_1\}$. 
Then we apply the CLIP image encoder to calculate the similarity between these two synthesized images and report the average similarity. 
As shown in the \textit{Swap}$^*$ case of Table~\ref{tab:swap}, our model presents stability to the mutation text order and arrives at the highest scores among others. 
\textbf{(2)}  We further investigate the model robustness against the order of words in the single-round generation. In particular, we randomly change the order of words in a sentence to generate the image. 
Then we calculate the CLIP score between generated images and ground-truth target images. As shown in the row \textit{Rotate}, our model also yields a competitive averaged similarity of $0.703$, which is higher than other methods. 
\textbf{(3)} We also mask out \textit{n} number of words in a sentence for the single-round generation. During training, we do not conduct such data augmentation, but we find that the model is still robust to such cases. 
As displayed in Table~\ref{tab:swap} case \textit{Mask n}, our model exceeds others in all conditions $n=2, 5, 9$. Even when more words are masked, the Clip Score of the proposed method only decreases slightly from $0.707$ to $0.692$.

\label{sec:lambda}
\noindent\textbf{Optimization on $\lambda$.} In Table~\ref{tab:lambda_m}, we study  different strategies on $\lambda$ during training. Since the final quality is mainly evaluated on the single-round generation, we have to find one trade-off between the image quality and the semantic alignment. We observe that gradually decreasing the weight of multi-round loss $\lambda$ is a simple yet effective way, which provides a trade-off between semantic alignment and generating quality.  

\begin{figure*}[t]
	\centering
 \vspace{.0in}
	\begin{overpic}[width=\textwidth]{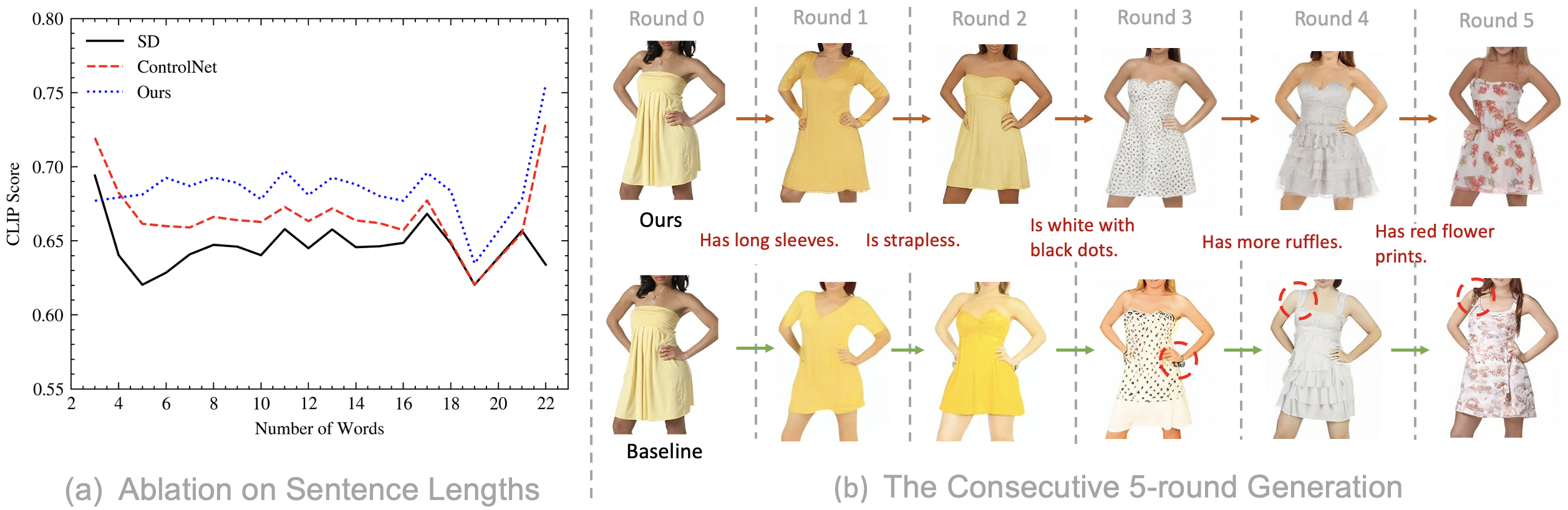}
        \end{overpic}
        \vspace{-.2in}
	\caption{
    (a) A comparative analysis of model performance across diverse text sentence lengths on FashionIQ for single-round generation. 
    (b) The consecutive 5-round generation. We could observe that the proposed method preserves superior generation consistency to the baseline, even after multiple rounds. 
	} \hspace{-3mm}\label{fig:sl_5rounds}
	\vspace{-.2in}
\end{figure*}

\noindent\textbf{Impact of the Sentence Length.} We further evaluate our model for different sentence lengths with CLIP Score (see Figure~\ref{fig:sl_5rounds}(a)). Since the proposed model usually learns the split long sentences, we observe a consistent improvement for both long sentences and short sentences. For extremely short sentences with less than 4 words, the proposed method still achieves a competitive performance. 

\noindent\textbf{Qualitative Results on Multiple Rounds.} Multiple-round editing ($\geq 3$ rounds) can be viewed as multiple 2-round generation. Therefore, we still could deploy the model trained on 2-round generation for multi-round editing. To explore the scalability of the 2-round model, we qualitatively compare our method in consecutive 5 rounds generation shown in Figure~\ref{fig:sl_5rounds}(b), where we can observe consistent attributes along consecutive rounds.


\section{Conclusion}
In this work, we introduce a new solution to the persistent challenge of fine-grained text-guided image generation. The proposed multi-round regularization, designed to maintain consistency across different modification orders, proves to be a consistent enhancement in addressing misalignment accumulation issues during interaction. 
Through extensive qualitative and quantitative evaluations, we show the high-fidelity generation quality achieved by our method, particularly in local modifications. 
Extending our evaluation to text-guided retrieval datasets such as FahisonIQ showcases the competitive performance of our approach in semantic alignment with text. 
Our work not only contributes to advancing the field of text-guided image generation but also opens avenues for more nuanced and reliable multi-round customization, addressing crucial details that are often overlooked in traditional single-round optimization approaches.

\newpage


{\small
\bibliographystyle{splncs04.bst}
\bibliography{egbib}
}

\end{document}